\pdfoutput=1

\documentclass[11pt]{article}

\usepackage{ACL2023}

\usepackage{times}
\usepackage{latexsym}

\usepackage[T1]{fontenc}

\usepackage[utf8]{inputenc}

\usepackage{microtype}

\usepackage{inconsolata}

\usepackage{hyperref}
\usepackage{url}

\usepackage{graphicx}
\usepackage{xspace,mfirstuc,tabulary}
\usepackage{times}
\usepackage{adjustbox}
\usepackage{comment}
\usepackage{latexsym}
\usepackage{booktabs}
\usepackage{multirow}
\usepackage{comment}
\usepackage{xspace}
\usepackage{color, colortbl}
\usepackage{xcolor}
\usepackage{enumitem}

\definecolor{Color}{gray}{0.9}

\newcommand*{\yoruba}{Yor\`ub\'a\xspace}
\newcommand*{\masakhanews}{\textbf{MasakhaNEWS} \xspace}

\title{MasakhaNEWS: News Topic Classification for African languages}

\author{\normalsize David Ifeoluwa Adelani$^{1*}$, Marek Masiak$^{1}$\thanks{\ \ \ Equal contribution} , Israel Abebe Azime$^{2}$, Jesujoba Oluwadara Alabi$^{2}$,  \\ 
\textbf{\normalsize Atnafu Lambebo Tonja$^{3,6}$,  
Christine Mwase$^{4}$, Odunayo Ogundepo$^{5}$, Bonaventure F. P. Dossou$^{6,7,8,9}$,} \\
\textbf{\normalsize Akintunde Oladipo$^{5}$, Doreen Nixdorf, Chris Chinenye Emezue$^{9,10}$, Sana Sabah al-azzawi$^{11}$,} \\
\textbf{\normalsize Blessing K. Sibanda, Davis David$^{12}$, Lolwethu Ndolela, Jonathan Mukiibi$^{13}$, Tunde Oluwaseyi Ajayi$^{14}$,} \\
\textbf{\normalsize Tatiana Moteu Ngoli$^{15}$, Brian Odhiambo, Abraham Toluwase Owodunni, Nnaemeka C. Obiefuna, } \\ 
\textbf{\normalsize Muhidin Mohamed$^{16}$, Shamsuddeen Hassan Muhammad$^{17}$, Teshome Mulugeta Ababu$^{18}$, } \\ 
\textbf{\normalsize Saheed Salahudeen Abdullahi$^{19}$, Mesay Gemeda Yigezu$^{3}$, Tajuddeen Gwadabe, Idris Abdulmumin$^{20}$,} \\
\textbf{\normalsize  Mahlet Taye Bame, Oluwabusayo Olufunke Awoyomi$^{21}$, 
 Iyanuoluwa Shode$^{22}$, Tolulope Anu Adelani, } \\
\textbf{\normalsize Habiba Abdulganiy Kailani, Abdul-Hakeem Omotayo$^{23}$, 
 Adetola Adeeko, Afolabi Abeeb, } \\
\textbf{\normalsize Anuoluwapo Aremu, Olanrewaju Samuel$^{24}$, Clemencia Siro$^{25}$,  Wangari Kimotho$^{26}$, } \\
\textbf{\normalsize Onyekachi Raphael Ogbu, Chinedu E. Mbonu$^{27}$, Chiamaka I. Chukwuneke$^{27,28}$,  Samuel Fanijo$^{29}$, } \\ 
\textbf{\normalsize Jessica Ojo,  Oyinkansola F. Awosan, Tadesse Kebede Guge$^{30}$, Sakayo Toadoum Sari$^{26,31}$, } \\ 
\textbf{\normalsize Pamela Nyatsine,  Freedmore Sidume$^{32}$, Oreen Yousuf, Mardiyyah Oduwole$^{33}$, Kanda P. Tshinu, } \\
\textbf{\normalsize Ussen Kimanuka$^{34}$,  Thina Diko, Siyanda Nxakama,  Sinodos Gebre$^{18}$, Abdulmejid Tuni Johar, 
} \\
\textbf{\normalsize Shafie Abdi Mohamed$^{34}$, Fuad Mire Hassan$^{35}$, Moges Ahmed Mehamed$^{36}$, Evrard Ngabire$^{37}$, } \\
\textbf{\normalsize  Jules Jules, Ivan Ssenkungu, and Pontus Stenetorp$^{1}$} \\ \\
\footnotesize
$^\forall$ Masakhane NLP, Africa, $^1$University College London, United Kingdom, $^2$ Saarland University, Germany,  \\
\footnotesize
$^3$ Instituto Politécnico Nacional, Mexico, $^4$Fudan University, China, $^5$ University of Waterloo, Canada,  $^{6}$ Lelapa AI, \\
\footnotesize
$^7$McGill University, Canada, $^8$ Mila Quebec AI Institute, Canada,  $^9$ Lanfrica,  $^{10}$ Technical University of Munich, Germany
 \\
\footnotesize
$^{11}$ Luleå University of Technology, Sweden, $^{12}$Tanzania Data Lab, Tanzania $^{13}$ Makerere University, Uganda, \\
\footnotesize
$^{14}$ Insight Centre for Data Analytics, Ireland,  $^{15}$Paderborn University, Germany,  $^{16}$Aston University, UK, \\
\footnotesize
 $^{17}$University of Porto, Portugal, $^{18}$ Dire Dawa University, Ethiopia $^{19}$Kaduna State University, Nigeria,  \\
\footnotesize
$^{20}$Ahmadu Bello University, Nigeria,  $^{21}$The College of Saint Rose, USA $^{22}$Montclair State University, USA, \\
\footnotesize
 $^{23}$ University of California, Davis, $^{24}$University of Rwanda, Rwanda 
 $^{25}$University of Amsterdam, The Netherlands, \\
\footnotesize
 $^{26}$AIMS, Cameroon, $^{27}$Nnamdi Azikiwe University, Nigeria , $^{28}$Lancaster University, United Kingdom,   \\
\footnotesize
$^{29}$Iowa State University, USA, $^{30}$Haramaya University, Ethiopia,   $^{31}$AIMS, Senegal, $^{32}$BIUST, Botswana,  \\ 
\footnotesize
$^{33}$NOUN, Nigeria, $^{34}$ PAUSTI, Kenya,  $^{34}$ Jamhuriya University, Somalia,  $^{35}$Somali National University,   \\
 \footnotesize
$^{36}$Wuhan University of Technology, China,  $^{37}$Deutschzentrum an der Universität Burundi,  Ethiopia\\  \\
\texttt{Correspondence: d.adelani@ucl.ac.uk}
}

\begin{document}
\maketitle
\begin{abstract}
Despite representing roughly a fifth of the world population, African languages are underrepresented in NLP research, in part due to a lack of datasets.
%
%
While there are individual language-specific datasets for several tasks, only a handful of tasks~(e.g. named entity recognition and machine translation) have datasets covering geographical and typologically-diverse African languages.
In this paper, we develop MasakhaNEWS---the largest dataset for news topic classification covering 16 languages widely spoken in Africa.
We provide and evaluate a set of baseline models by training classical machine learning models and fine-tuning several language models.
Furthermore, we explore several alternatives to full fine-tuning of language models that are better suited for zero-shot and few-shot learning, such as: cross-lingual parameter-efficient fine-tuning~(MAD-X), pattern exploiting training~(PET), prompting language models~(ChatGPT), and prompt-free sentence transformer fine-tuning~(SetFit and the co:here embedding API).
%
%
Our evaluation in a few-shot setting, shows that with as little as 10 examples per label, we achieve more than 90\%~(i.e. 86.0 F1 points) of the performance of fully supervised training~(92.6 F1 points) leveraging the PET approach.
Our work shows that existing supervised approaches work well for all African languages and that language models with only a few supervised samples can reach competitive performance, both findings which demonstrate the applicability of existing NLP techniques for African languages.
\end{abstract}

\section{Introduction}
News topic classification is a 
text classification task in NLP that involves categorizing news articles into different categories like sports, business, entertainment, and politics. 
It has shaped the development of several machine learning algorithms over the years, such as topic modeling~\citep{lda_blei,dieng-etal-2020-topic} and deep learning models~\citep{NIPS2015_zhang,joulin-etal-2017-bag}. Similarly, news topic classification is a popular downstream task for evaluating the performance of large language models~(LLMs) for both fine-tuning and prompt-tuning setups~\citep{Yang2019XLNetGA,Sun2019HowTF,NEURIPS2020_gpt3,Pengfei_prompt}.  

Despite the popularity of the task in bench-marking LMs, most of the evaluation have only been performed on English and a few other high-resource languages.
It is \textit{unclear how this approach extends to pre-trained multilingual language models}  for low-resource languages.
For instance, BLOOM~\citep{Scao2022BLOOMA1} 
was pre-trained on 46 languages, including 22 African languages~(mostly from  the Niger-Congo family).
However, extensive evaluation on these set of African languages was not performed due to lack of evaluation datasets. In general, only a handful of NLP
tasks such as machine translation~\citep{adelani-etal-2022-thousand, team2022NoLL}, named entity recognition~\citep{adelani2021masakhaner,adelani2022masakhaner}, and sentiment classification~\citep{muhammad2023afrisenti} have standardized benchmark datasets covering several geographical and typologically-diverse
African languages. 
Another popular task that can be used for evaluating the downstream performance of language models is news topic classification, but human-annotated datasets for benchmarking topic classification using language models for African languages are \textit{scarce}. 


In this paper, we address two problems: the lack of evaluation datasets and lack of extensive evaluation of LMs for African languages. We create a large-scale \textbf{news topic classification} dataset covering 16 typologically-diverse languages widely spoken in Africa, including English and French, with the same label categories across all languages. Our dataset is also suitable for \textbf{news headline generation} task~\citep{Aralikatte2023VrtaAL}: a special type of text summarization. 
We provide several baseline models using both classical machine learning approaches and fine-tuning LMs. 
Furthermore, we explore several alternatives to full fine-tuning of language models that are better suited for zero-shot and few-shot learning~(e.g. 5-examples per label) such as cross-lingual parameter-efficient fine-tuning~(MAD-X~\citep{pfeiffer-etal-2020-mad}), pattern exploiting training~(PET)~\citep{schick-schutze-2021-exploiting}, prompting ChatGPT LLM, and prompt-free, sentence transformer fine-tuning~(SetFit)~\citep{Tunstall2022EfficientFL}, and the co:here embedding API. 

Our evaluation in a zero-shot setting shows the potential of prompting ChatGPT for news topic classification for low-resource African languages. We found that GPT-3.5-Turbo has impressive result for languages that make use of Latin script, but \textit{perform poorly for non-Latin based scripts like Amharic and Tigrinya}. However, \textit{GPT-4 was able to overcome this challenge for non-Latin script} with impressive performance matching the result of cross-lingual transfer experiments from a related African language. 

In a few-shot setting, we show that with as little as 10 examples per label, we achieved more than 90\%~(i.e. 86.0 F1 points) of the performance of full supervised training~(92.6 F1 points) leveraging the PET approach. We hope this encourages the NLP community to benchmark and evaluate LLMs on more low-resource languages. For reproducibility, we release our data and code under academic license or CC BY-NC 4.0 on Github.\footnote{\url{https://github.com/masakhane-io/masakhane-news}} 

\section{Related Work}
\label{gen_inst}
\paragraph{News topic classification}, an application of text classification, is a popular task in natural language processing. There are various news topic classification datasets, including BBC News~\citep{greene_bbc}, AG News~\citep{NIPS2015_zhang}, and the multimodal N24News~\citep{wang-etal-2022-n24news}, all of which are English datasets. In addition, there is the IndicNLP News~\citep{kunchukuttan2020indicnlpcorpus} which is a multilingual dataset for Indian langauges. For African languages, only a handful of human annotated datasets exists, such as the Hausa \& \yoruba dataset~\citep{hedderich-etal-2020-transfer} (only covering news headline), KINNEWS \& KIRNEWS datasets for Kinyarwanda and Kirundi~\citep{niyongabo-etal-2020-kinnews}, and Tigrinya News~\citep{info12020052}. Others are semi-automatically created using predefined topics from news websites like Amharic news~\citep{DBLP:journals/corr/abs-2103-05639} and ANTC dataset~\citep{alabi-etal-2022-adapting}---that covered five African languages (Lingala, Somali, Naija, Malagasy, and isiZulu). These datasets, however, have limitations due to the fact that they were created with little or no human supervision and using different labeling schemes. In contrast, in this work we present news topic classification data for 16 typologically diverse African languages with a consistent labeling scheme across all languages. 


\smallskip \noindent \textbf{Prompting Language Models} using manually designed prompts to guide text generation has recently been applied to a myriad of NLP tasks, including topic classification. Models such as GPT-3~\citep{NEURIPS2020_gpt3} and T5~\citep{2020t5,sanh2022multitask} are able to learn more structural and semantic relationships between words and have shown impressive results even in multilingual scenarios when tuned for different tasks~\citep{chung-etal-2022-flant5,muennighoff-etal-2023-crosslingual}.
One approach to prompt-tuning a language model for topic classification is to design a ``template'' for classification and insert a sequence of text into template~\citep{gao-etal-2021-making,shin-etal-2020-autoprompt}. 

There are some other 
approaches to few-shot learning without prompting.
One of them is SetFit~\citep{Tunstall2022EfficientFL}, which takes advantage of sentence transformers to generate dense representations for input sequences.
These representations are then passed through a classifier to predict class labels.
The sentence transformers are trained on a few examples using contrastive learning where positive and negative training pairs are sampled by in-class and out-class sampling.
Another common approach is Pattern-Exploiting Training also known as PET~\citep{schick-schutze-2021-exploiting}. PET is a semi-supervised training approach that used restructured input sequences to condition language models to better understand a given task, while iPET~\citep{schick-schutze-2021-just} is an iterative variant of PET that is also shown to perform better. 



\section{Languages}
\label{languages}

\begin{table*}[t]

\begin{center}
\scalebox{0.68}{
\begin{tabular}{lllr|p{30mm}r}
\textbf{Language}
& \textbf{Family/branch}
& \textbf{Region}
& \textbf{\# speakers} & \textbf{News Source} & \textbf{\# articles} \\
\midrule
Amharic (amh)  & Afro-Asiatic / Ethio-Semitic & East Africa    & 57M & BBC & 8,204 \\
English (eng) & Indo-European / Germanic & Across Africa   & 1268M & BBC & 5,073  \\
French (fra) & Indo-European /Romance & Across Africa   & 277M & BBC & 5,683 \\
Hausa (hau) & Afro-Asiatic / Chadic & West Africa    & 77M & BBC & 6,965 \\
Igbo (ibo) & Niger-Congo / Volta-Niger & West Africa    & 31M  &  BBC & 4,628 \\
Lingala (lin) &  Niger-Congo / Bantu & Central Africa    & 40M & VOA & 2,022\\
Luganda (lug) &  Niger-Congo / Bantu & Central Africa    & 11M & Gambuuze & 2,621 \\
Naija (pcm)  & English Creole  & West Africa    & 121M & BBC & 7,783 \\
Oromo (orm) & Afro-Asiatic / Cushitic & East Africa  & 37M & BBC & 7,782 \\
Rundi (run) & Niger-Congo / Bantu & East Africa & 11M & BBC & 2,995 \\
chiShona (sna) & Niger-Congo / Bantu & Southern Africa   & 11M & VOA \& Kwayedza & 11,146 \\
Somali (som) & Afro-Asiatic / Cushitic & East Africa   & 22M & BBC  & 2,915 \\
Kiswahili (swa) & Niger-Congo / Bantu & East \& Central Africa  & 71M-106M &  BBC & 6,431 \\
Tigrinya (tig) & Afro-Asiatic / Ethio-Semitic & East Africa   & 9M & BBC & 4,372 \\
isiXhosa (xho) & Niger-Congo / Bantu & Southern Africa  & 19M & Isolezwe & 24,658  \\
\yoruba (yor) & Niger-Congo / Volta-Niger & West Africa   & 46M & BBC & 6,974 \\
\bottomrule
\end{tabular}
}
\vspace{-2mm}
\caption{\textbf{Languages covered in  and Data Source}: including language family, region, number of  L1 \& L2 speakers, and number of articles from each news source.}
\label{tab:languages}

\end{center}
\vspace{-4mm}
\end{table*}
~\autoref{tab:languages} presents the languages covered in  along with information on their language families, their primary geographic regions in Africa, and the number of speakers. Our dataset consists of a total of 16 typologically-diverse languages, and they were selected based on the availability of publicly available news corpora in each language, the availability of native-speaking annotators, geographical diversity and most importantly, because they are widely spoken in Africa. 
English and French are official languages in 42 African countries, Swahili is native to 12  countries, and Hausa is native to 6 countries. In terms of geographical diversity, we have four languages spoken in West Africa, seven languages spoken in East Africa, two languages spoken in Central Africa~(i.e. Lingala and Kiswahili), and two spoken in Southern Africa~(i.e chiShona and isiXhosa). Also, we cover four language families, Niger-Congo~(8) Afro-Asiatic~(5), Indo-European~(2), and English Creole~(1). The only English creole language is Nigerian-Pidgin, also known as Naija. Each language is spoken by at least 10 million people, according to Ehnologue~\citep{ethnologue}. 
\vspace{3mm}



\section{\masakhanews dataset}
\label{data}

\subsection{Data Source}
The data used in this research study were sourced from multiple reputable news outlets. The collection process involved crawling the British Broadcasting Corporation~(BBC) and Voice of America~(VOA) websites.  
We crawled between 2k--12k articles depending on the number of articles available on the websites. Some of the websites already have some pre-defined categories, we make use of this to additionally filter articles that do not belong to categories we plan to annotate. We took \textit{inspiration} of news categorization from \textbf{BBC English} with six~(6) pre-defined and well-defined categories \textit{(``business'', ``entertainment'', ``health'', ``politics'', ``sports'', and ``technology'')} with over 500 articles in each category. For English, we only crawled articles belonging to these categories while for the other languages, we crawled all articles. Our target is to have around \textbf{3,000} articles for annotation but three languages~(Lingala, Rundi, and Somali) have less than that. \autoref{tab:dataset} shows the news source per language and the number of articles crawled.

\subsection{Data Annotation}

We recruited volunteers from the Masakhane community---an African grassroots community focused on advancing NLP for African languages.\footnote{all annotators are were included as authors of the paper.} The annotators were asked to label 3k articles into eight categories: ``\textit{business}'', ``\textit{entertainment}'', ``\textit{health}'', ``\textit{politics}'', ``\textit{religion}'', ``\textit{sports}'', ``\textit{technology}'', and ``\textit{uncategorized}''. Six of the categories are based on BBC English major news categories, the ``\textit{religion}'' label was added since many African news websites frequently cover this topic. Other articles that do not belong to the first seven categories, are assigned to the ``\textit{uncategorized}'' label. 

For each language, the annotation followed two stages. In the \textbf{first stage}, we randomly shuffled the entire dataset and asked annotators to label the first 200 articles manually.  In the \textbf{second stage}, we made use of active learning by combining the first 200 annotated articles with articles with pre-defined labels 
where available, and trained a classifier~(i.e. by fine-tuning AfroXLMR-base~\citep{alabi-etal-2022-adapting}). We ran predictions on the rest of the articles, and asked annotators to correct the mistakes of the classifier. This approach helped to speed up the annotation process.

\begin{table*}[t] \centering
\scalebox{0.7}{
\begin{tabular}{lrr|rrrrrrr|rr}
\toprule
 & & \multicolumn{7}{c}{\textbf{Topics (number of articles per topic)}} & & & \textbf{Fleiss}  \\ 
\textbf{Language}  & \textbf{Train/Dev/Test} & \textbf{\# topics}  & \textbf{\# bus} & \textbf{\# ent} & \textbf{\# health} & \textbf{\# pol} & \textbf{\# rel} & \textbf{\# sport} & \textbf{\# tech} &\textbf{\# Annotator} & \textbf{Kappa} \\ \midrule
Amharic (amh)  & 1311/ 188/ 376 & 4 & 404 & - & 500 & 500 & - & 471 & - & 5 & 0.81 \\ 
English (eng) &  3309/ 472/ 948  & 6 & 799 & 750 & 746 & 821 & - & 1000 & 613 & 7 & 0.81 \\ 
French (fra)  & 1476/ 211/ 422   & 5 & 500 & - & 500 &  500 & - & 500 & 109 & 3 & 0.83 \\
Hausa (hau)  & 2219/ 317/ 637  & 7 & 399 & 500 & 493 & 500 & 493 &  497 & 291 & 5 & 0.85 \\ 
Igbo (ibo)  & 1356/ 194/ 390  & 6 & 292 & 366 & 424 & 500 & 73 & 285 & - & 4 & 0.65\\ 
Lingala (lin) & 608/ 87/ 175 & 4 & 82 & - &  193 & 500 & - & 95 & - & 2 & 0.56\\
Luganda (lug) & 771/ 110/ 223 & 5 & 169 & - &  228 & 500 & 91 & 116 & - & 1 & - \\ 
Oromo (orm)  & 1015/ 145/ 292 & 4 & - & 119 & 447 & 500 & - & 386 & - & 3 & 0.63 \\ 
Naija (pcm)   & 1060/ 152/ 305 & 5 & 97 & 460 & 159 & 309 & - & 492 & - & 4 & 0.66 \\ 
Rundi (run)   & 1117/ 159/ 322 & 6 & 76 & 158 & 372 & 500 & 73 & 419 & - & 1  & - \\ 
chiShona (sna)   & 1288/ 185/ 369 & 4 & 500 & - &  425 & 500 & - & 417 & -  & 3 & 0.63 \\ 
Somali (som)   & 1021/ 148/ 294 & 7 & 114 & 139 & 354 & 500 & 73 & 148 & 135 & 3 & 0.55 \\ 
Kiswahili (swa)  & 1658/ 237/ 476  & 7 & 316 & 98 & 500  & 500 & 292 & 500 & 165 & 4 & 0.72 \\ 
Tigrinya (tir)   & 947/ 137/ 272 & 6 & 80 & 167 & 395 & 500 & - & 125 & 89 & 2 & 0.63\\
isiXhosa (xho)   & 1032/ 147/ 297 & 5 &  72 & 500 & 100 & 308 & - & 496 & - & 3 & 0.89\\ 
\yoruba (yor)  & 1433/ 206/ 411 & 5 & -  & 500 & 398 & 500 & 317 & 335 &  -  & 5 & 0.80 \\
\bottomrule
\end{tabular}
}
\vspace{-2mm}
\caption{\textbf{\masakhanews dataset}. The size of the annotated data, news topics, and number of annotators.  Topics are labelled by their prefixes in the table~(\textbf{topics}):  \textbf{bus}iness, \textbf{ent}ertainment, \textbf{health}, \textbf{pol}itics, \textbf{rel}igion, \textbf{sport}, \textbf{tech}nology. }
\label{tab:dataset}

\end{table*}

\paragraph{Annotation tool} We make use of an in-house annotation tool 
to label the articles.  \autoref{sec:tool} shows an example of the interface of the tool. To further simplify the annotator effort, we ask annotators to label articles based on the headlines instead of the entire article. However, since some headlines are not very descriptive, we decided to concatenate the headline and the first two sentences of the news text to provide additional context to annotators. 

\paragraph{Inter-agreement score} We report Fleiss Kappa score~\citep{fleiss1971mns} to measure the agreement of annotation. \autoref{tab:dataset} shows that all languages have a moderate to perfect Fleiss Kappa score~(i.e. 0.55 - 0.85), which shows a high agreement among the annotators recruited for each language. Languages with only one annotator~(i.e. Luganda and Rundi) were excluded in the evaluation. 

\paragraph{Deciding a single label per article} After annotation, we assigned the final label to each article by majority voting. Each label of an article needs to be agreed by a minimum of two annotators to be assigned the label. We only had exceptions for Luganda and Rundi, since they had one annotator. Our final dataset for each language consist of a minimum of 72 articles per topic, and a maximum of 500, except for English language where the classes are roughly balanced. We excluded the infrequent labels so we do not have a highly unbalanced dataset. The choice of a minimum of 72 articles ensures a minimum of 50 articles in the training set.~\footnote{since we require 50 instances per class or 50-shots for the few-shot experiments in (\S\ref{sec:few_shot_eval})} Our target is to have at least four topics per language with a minimum of 72 articles. This approach worked smoothly except for two languages: Lingala~(``politics'', ``health'' and ``sports'') and chiShona~(``business'', ``health'' and ``politics''), where we had only three topics with more than 72 articles. To ensure we have more articles per class, we had to resolve the conflict in annotation between Lingala annotators to ensure we have more labels for the ``business'' category. This approach still results in infrequent classes for chiShona. We had to crawl additional ``sports'' articles from a local chiShona website~(\textit{Kwayedza}), followed by manual filtering of unrelated sports news.

\paragraph{Data Split} \autoref{tab:dataset} provides the data split for  languages. We also provide the distribution of articles by topics. We divided the annotated data into TRAIN, DEV and TEST split following 70\% / 10\% / 20\% split ratio.





\section{Baseline Experiments}
\label{headings}
We trained baseline text classification models by concatenating the news headline and news text using different approaches. 





\subsection{Baseline Models}

\begin{table*}[t]
\begin{center}
\footnotesize
\begin{adjustbox}{width=\textwidth,center}

\begin{tabular}{lcrrrrrrrrrrrrrrrr|c}
\toprule
\textbf{Model} & \textbf{size} & \textbf{amh} & \textbf{eng} & \textbf{fra} & \textbf{hau} & \textbf{ibo} & \textbf{lin} & \textbf{lug} & \textbf{orm} & \textbf{pcm} & \textbf{run} & \textbf{sna} & \textbf{som} & \textbf{swa} & \textbf{tir} & \textbf{xho} & \textbf{yor} & \textbf{AVG}  \\
\midrule
\multicolumn{7}{l}{\textit{classical ML}} \\
MLP & $<$20K & 92.0 & 88.2 & 84.6 & 86.7 & 80.1 & 84.3 & 82.2 & 86.7 & 93.5 & 85.9 & 92.6 & 71.1 & 77.9 & 81.9 & 94.5 & 89.3 & 85.7 \\
NaiveBayes & $<$20K  & 91.8 & 83.7 & 84.3 & 85.3 & 79.8 & 82.8 & 84.0 & 85.6 & 92.8 & 79.9 & 91.5 & 74.8 & 76.6 & 71.4 & 91.0 & 84.0 & 83.7 \\
XGBoost & $<$20K & 90.1 & 86.0 & 81.2 & 84.7 & 78.6 & 74.8 & 83.8 & 83.2 & 93.3 & 79.2 & 94.3 & 68.5 & 74.9 & 75.2 & 91.1 & 85.2 & 82.8  \\
\midrule
\multicolumn{7}{l}{\textit{multilingual text encoders}} \\
\rowcolor{Color}
AfriBERTa & 126M & 90.6 & 88.9 & 76.4 & 89.2 & 87.3 & 87.0 & 85.1 & 89.4 & 98.1 & 91.3 & 89.3 & 83.9 & 83.3 & 87.0 & 86.9 & 90.3 & 87.8 \\
XLM-R-base & 270M & 90.9 & 90.6 & 90.4 & 88.4 & 82.5 & 87.9 & 65.3 & 82.2 & 97.8 & 85.9 & 88.9 & 73.8 & 85.6 & 54.6 & 78.6 & 84.5 & 83.0 \\
\rowcolor{Color}
AfroXLMR-base & 270M & 94.2 & 92.2 & \textbf{92.5} & 91.0 & 90.7 & 93.0 & 89.4 & 92.1 & 98.2 & 91.4 & 95.4 & 85.2 & 88.2 & 86.5 & 94.7 & 93.0 & 91.7 \\
\rowcolor{Color}
AfroLM & 270M & 90.3 & 87.7 & 77.5 & 88.3 & 85.4 & 85.7 & 88.0 & 83.5 & 95.9 & 86.8 & 92.5 & 72.0 & 83.2 & 83.5 & 91.4 & 86.5 & 86.1 \\
mDeBERTa & 276M & 91.7 & 90.8 & 89.2 & 88.6 & 88.3 & 81.6 & 65.7 & 84.7 & 96.8 & 89.4 & 93.9 & 72.0 & 84.6 & 78.7 & 90.5 & 89.3 & 86.0 \\
LABSE & 471M & 92.5 & 91.6 & 90.9 & 90.0 & 91.6 & 89.6 & 86.8 & 86.7 & 98.4 & 91.1 & 94.6 & 82.1 & 87.6 & 83.8 & 94.7 & 92.1 & 90.3 \\
XLM-R-large & 550M & 93.1 & 92.2 & 91.4 & 90.6 & 84.2 & 91.8 & 73.9 & 88.4 & 98.4 & 87.0 & 88.9 & 76.1 & 85.6 & 62.7 & 89.2 & 84.5 & 86.1   \\
\rowcolor{Color}
AfroXLMR-large & 550M & \textbf{94.4} &\textbf{93.1} & 91.1 & \textbf{92.2} & \textbf{93.4} & \textbf{93.7} & \textbf{89.9} & \textbf{92.1} & \textbf{98.8} & \textbf{92.7} & \textbf{95.4} & \textbf{86.9} &\textbf{87.7} & \textbf{89.5} & \textbf{97.3} & \textbf{94.0} & \textbf{92.6}\\
RemBERT & 559M  & 92.4 & 92.4 & 90.8 & 90.5 & 91.1 & 91.5 & 86.7 & 88.7 & 98.2 & 90.6 & 93.9 & 75.9 & 86.7 & 69.9 & 92.5 & 93.0 & 89.1 \\
\midrule
\multicolumn{7}{l}{\textit{multilingual text-to-text LMs}} \\
\rowcolor{Color}
AfriTeVa-base & 229M & 87.0 & 80.3 & 71.9 & 85.8 & 79.9 & 82.8 & 60.2 & 82.9 & 95.2 & 80.0 & 84.4 & 58.0 & 80.7 & 55.2 & 69.4 & 86.4 & 77.5  \\
mT5-base & 580M & 78.2 & 89.8 & 59.0 & 82.7 & 76.8 & 80.8 & 75.0 & 79.2 & 96.1 & 85.7 & 90.4 & 75.0 & 76.1 & 65.1 & 71.8 & 86.2 & 80.0 \\
Flan-T5-base & 580M &  54.5 & 92.4 & 88.9 & 84.5 & 86.6 & 90.6 & 84.1 & 85.8 & 97.8 & 87.3 & 90.6 & 76.0 & 79.0 & 41.5 & 90.8 & 88.0 & 82.4   \\
\rowcolor{Color}
AfriMT5-base & 580M & 90.2 & 90.3 & 87.4 & 87.9 & 88.0 & 88.6 & 84.8 & 83.9 & 96.6 & 91.0 & 91.5 & 77.8 & 84.4 & 80.8 & 91.6 & 88.8 & 87.7 \\



\bottomrule
    \end{tabular}
\end{adjustbox}
\caption{\textbf{Baseline results on }. We compare several ML approaches using both classical ML and LMs. Average is over 5 runs.  Evaluation is based on weighted F1-score. Africa-centric models are in gray color }
\label{tab:masakhanews_baselines}
  \end{center}
  \vspace{-4mm}
\end{table*}

We trained three classical ML models: Naive Bayes, multi-layer perceptron, and XGBoost using the popular \texttt{sklearn} tool~\citep{scikit-learn}.  We employed the ``CountVectorizer'' method to represent the text data, which converts a collection of text documents to a matrix of token counts. This method allows us to convert text data into numerical feature vectors. 

Furthermore, we fine-tune nine kinds of multilingual text encoders, seven of them are BERT/RoBERTa-based i.e. XLM-R~(base \& large)~\citep{conneau-etal-2020-unsupervised}, AfriBERTa-large~\citep{ogueji-etal-2021-small}, RemBERT~\citep{chung2021rethinking}, AfroXLMR (base \& large)~\citep{alabi-etal-2022-adapting}, and AfroLM~\citep{dossou2022afrolm}, the other two are 
mDeBERTaV3~\citep{He2021DeBERTaV3ID}, and LaBSE~\citep{feng-etal-2022-language}. mDeBERTaV3 pre-trained a DeBERTa-style model~\citep{he2021deberta} with replaced token detection objective proposed in ELECTRA~\citep{clark2020electra}. On the other hand, LaBSE is a multilingual sentence transformer model that is popular for mining parallel corpus for machine translation. 

Finally, we fine-tuned four multilingual Text-to-Text (T2T) models, mT5-base~\citep{xue-etal-2021-mt5}, 
Flan-T5-base~\citep{chung-etal-2022-flant5}, AfriMT5-base~\citep{adelani-etal-2022-thousand}, AfriTeVA-base~\citep{jude-ogundepo-etal-2022-afriteva}. 
The fine-tuning and evaluation of the multilingual text-encoders and T2T models were performed using HuggingFace Transformers~\citep{wolf-etal-2020-transformers} and PyTorch Lightning\footnote{\url{https://pypi.org/project/pytorch-lightning/}}. The models were fine-tuned on Nvidia V100 GPU for 20 epochs, batch size of 32, $1e-5/5e-5$ lr, and max. sequence length of 256. 

The LMs evaluated were both massively multilingual (i.e. typically trained on over 100 languages around the world) and African-centric (i.e. trained mostly on languages spoken in Africa). 
The African-centric multilinual text encoders are all modeled after XLM-R. AfriBERTa was pretrained from scratch on 11 African languages, AfroXLMR was adapted to African languages through fine-tuning the original XLM-R model on 17 African languages and 3 languages commonly spoken in Africa, while AfroLM was pretrained on 23 African languages utilizing active learning. Similar to the multilingual text encoders, the T2T models used in this study were pretrained on hundreds of languages, and they are all based on the T5 model~\citep{2020t5}, which is an encoder-decoder model trained with the span-mask denoising objective. mT5 is a multilingual version of T5, 
and Flan-T5 was fine-tuned on multiple tasks using T5 as a base. The study also included adaptations of the original models, such as AfriMT5-base, as well as AfriTeVA-base, a T5 model pre-trained on 10 African languages.

\begin{figure}[t]
    \centering
   \includegraphics[width=\linewidth]{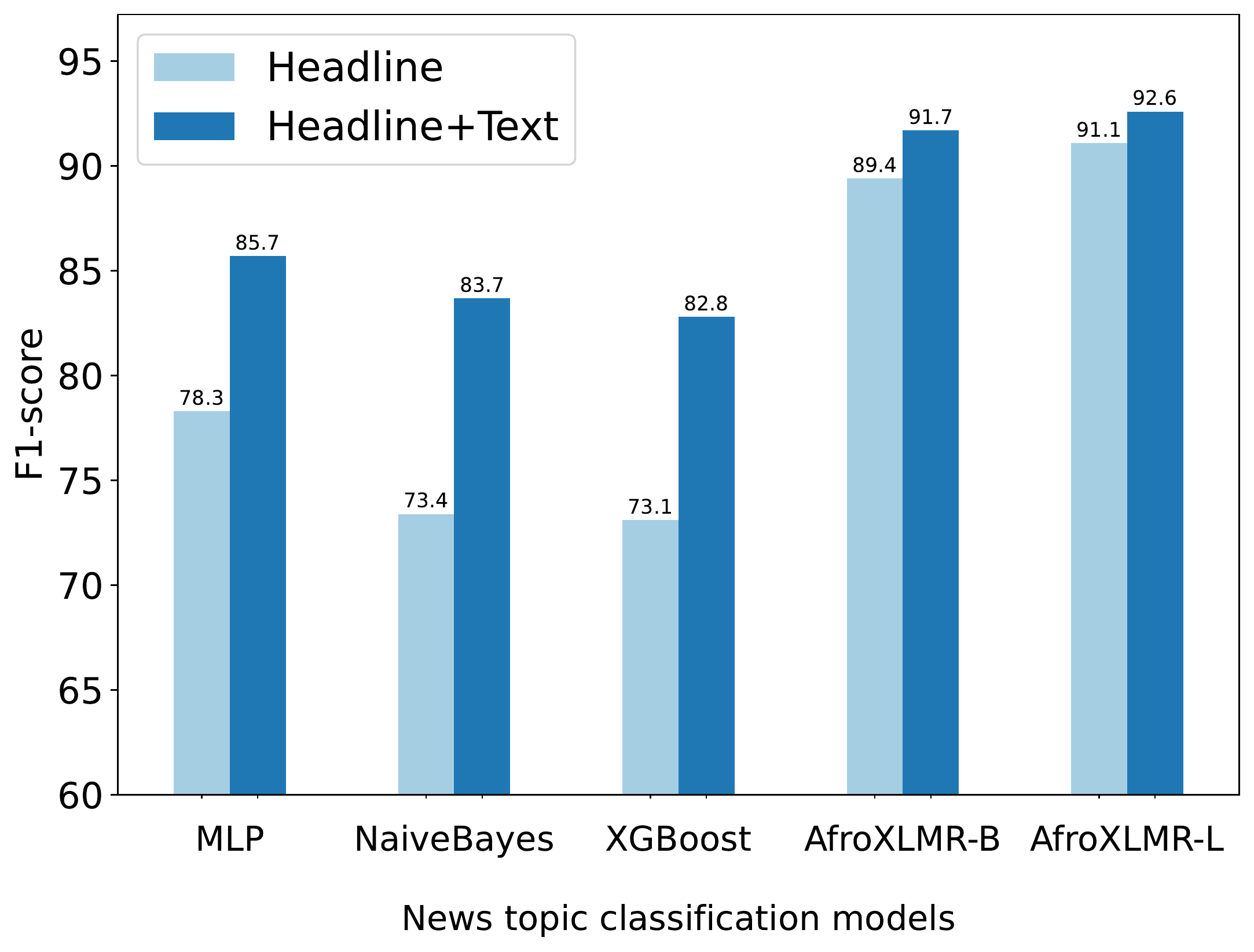}
    \caption{\textbf{Comparison of article content type used for training news topic classification models}. We report the average  across all languages when either  \texttt{headline} or \texttt{headline+text} is used}
    \label{fig:comparison_headline}
\end{figure}

\subsection{Baseline Results}
\autoref{tab:masakhanews_baselines} shows the result of training several models on  TRAIN split and evaluation on the TEST split for each language. 
Our evaluation shows that classical ML models are worse in general than fine-tuning multilingual LMs  on average, however, the drop in performance is sometimes comparable to LMs if the language was not covered during the pre-training. 
For example, MLP, NaiveBayes and XGBoost have better performance than AfriBERTa on \texttt{fra} and \texttt{sna} since they were not seen during pre-training of the LM. 
Similarly, AfroLM had worse result for \texttt{fra} for the same reason. 
On average, XLM-R-base, AfroLM, mDeBERTaV3, XLM-R-large  gave $83.0$ F1, $86.1$ F1, $86.0$ F1, and $86.1$ F1 respectively, with worse performance compared to the other LMs ($87.8-92.6$ F1) because they do not cover some of the African languages during pre-training (see \autoref{tab:plm_languages}) or they have been pre-trained on a small data (e.g. AfroLM pretrained on less than 0.8GB despite seeing 23 African languages during pre-training). 
Larger models such as LABSE and RemBERT that cover more languages performed better than the smaller models, for example, LABSE achieved over of $2.5$ F1 points over AfriBERTa.  

The best result achieved is by AfroXLMR-base/large with over $4.0$ F1 improvement over AfriBERTa. The larger variant gave the overall best result due to the size. AfroXLMR models benefited from being pre-trained on most of the languages we evaluated on. We also tried multilingual T2T models, but none of the models reach the performance of AfroXLMR-large despite their larger sizes. We  observe the same trend that the adapted mT5 model (i.e. AfriMT5) gave better result compared to mT5 similar to how AfroXLMR gave better result than XLM-R. We found \textit{FlanT5-base to be competitive to AfriMT5} despite seeing few African languages, however, \textit{the performance was very low for languages that uses the Ge'ez script} like \texttt{amh} and \texttt{tir} since the model do not support Ge'ez.

\paragraph{Headline-only training} We compare our results using \texttt{headline+text} (as shown in \autoref{tab:masakhanews_baselines}) with training on the article \texttt{headline}---with shorter content, we find out that fine-tuned LMs gave impressive performance with only headlines while classical ML methods struggle due to shorter content. \autoref{fig:comparison_headline} shows the result of our comparison. AfroXLMR-base and AfroXLMR-large both improve by ($2.3$) and  ($1.5$) F1 points respectively if we use \texttt{headline+text} instead of \texttt{headline}. Classical ML models improve the most when we make use of \texttt{headline+text} instead of \texttt{headline}; MLP, NaiveBayes and XGBoost improve by large F1 points (i.e. $7.4-9.7$). Thus, for the remainder of this paper, we make use of \texttt{headline+text}. \autoref{sec:article_type} provides the breakdown of the result by languages for the comparison of  \texttt{headline} and  \texttt{headline+text}.


\begin{table*}[t]
\begin{center}
\footnotesize
\begin{adjustbox}{width=\textwidth,center}
\begin{tabular}{lrrrrrrrrrrrrrrrr|cc}
\toprule
\textbf{SRC LANG} &  \textbf{amh} & \textbf{eng} & \textbf{fra} & \textbf{hau} & \textbf{ibo} & \textbf{lin} & \textbf{lug} & \textbf{orm} & \textbf{pcm} & \textbf{run} & \textbf{sna} & \textbf{som} &\textbf{swa} & \textbf{tir} & \textbf{xho} & \textbf{yor} & \textbf{AVG} & \textbf{AVG$^{src}$}  \\
\midrule
\multicolumn{7}{l}{\textit{Fine-tune (AfroXLMR-base)}} \\
\texttt{hau} & 81.8 & 78.8 & 72.9 & 91.5 & 83.2 & 74.4 & 57.5 & 63.3 & 93.2 & 81.6 & 85.5 & 63.3 & 80.7 &  73.2 & 77.4 & 80.4 & 77.4 & 76.2 \\

\texttt{swa}  & 89.5 & 82.4 & 86.7 & 80.8 & 81.5 & 74.5 & 66.5 & 63.8 & 92.7 & 86.2 & 83.6 & 74.7 & 87.3 & 71.8 & 72.6 & 80.4 & 79.7 & 79.1 \\
\midrule
\multicolumn{7}{l}{\textit{MAD-X}} \\
\texttt{hau} & 81.0 & 79.5 & 72.2 & 90.3 & 87.4 & 82.6 & 84.4 & 80.2 & 91.2 & 76.0 & 89.9 & 66.5 & 81.2 & 72.6 & 82.8 & 87.4 & 81.6 & 81.0 \\

\texttt{swa}  & 91.0 & 80.9 & 86.1 & 81.2 & 83.0 & 85.0 & 75.1 & 82.6 & 94.2 & 86.9 & 90.1 & 74.6 & 88.4 & 77.6 & 80.7 & 88.8 & \textbf{84.1} & \textbf{84.0}\\
\midrule

\multicolumn{7}{l}{\textit{PET}} \\
None   &  67.2 & 53.3 & 51.7 & 42.1 & 50.4 & 28.6 & 27.0 & 43.9 & 63.1 & 57.9 & 62.2 & 39.2 & 53.8 & 45.2 & 56.0 & 49.7 & 49.5 & 49.7\\
\midrule
\multicolumn{7}{l}{\textit{SETFIT}} \\
None   &  75.8 & 61.6 & 60.1 & 53.3 & 53.1 & 59.6 & 40.1 & 38.9 & 72.0 & 55.1 & 66.6 & 49.4 & 55.2 & 37.8 & 49.3 & 63.7 & 55.7 & 55.9\\
\midrule
\multicolumn{7}{l}{\textit{ChatGPT (GPT 3.5 Turbo) - Mar 23 version}} \\
None   &  33.3 & 79.3 & 67.6 & 59.4 & 65.0 & 62.3 & 59.4 & 62.9 & 93.2 & 73.6 & 73.0 & 62.0 & 69.3 & 41.4 & 73.9 & 80.1 & 66.0 & 66.2 \\
\midrule
\multicolumn{7}{l}{\textit{ChatGPT (GPT 3.5 Turbo) - May 24 version}} \\
None   &  36.1 & 79.5 & 69.6 & 70.1 & 78.3 & 75.1 & 64.7 & 72.0 & 93.1 & 82.2 & 84.5 & 72.3 & 75.9 & 45.0 & 78.0 & 81.7 & 72.4 & 72.3 \\
\midrule
\multicolumn{7}{l}{\textit{GPT 4} -- May 24 version} \\
None   &  88.5 & 79.1 & 77.3 & 76.5 & 84.0 & 82.6 & 77.9 & 70.0 & \textbf{96.2} & \textbf{88.6} & \textbf{90.8} & \textbf{77.3} & 75.0 & 76.7 & \textbf{83.1} & 83.7 & 81.7 & 82.5 \\
\bottomrule
    \end{tabular}
\end{adjustbox}
\vspace{-2mm}
\caption{\textbf{Zero-shot learning on }. We compare several approaches such as using MAD-X, PET and SetFit. We excluded the source languages \texttt{hau} and \texttt{swa} from the average (AVG$^{src}$).} 
\label{tab:zeroshot_result}
  \end{center}
\end{table*}

\section{Zero-shot and Few-shot transfer}

\subsection{Methods}
Here, we compare different zero-shot and few-shot methods: 
\paragraph{Fine-tune} (Fine-tune on a \textit{source language}, and evaluate on a \textit{target language}) using AfroXLMR-base. This is only used in the \textbf{zero-shot setting}. 

\paragraph{MAD-X}~\citep{pfeiffer-etal-2020-mad,pfeiffer-etal-2021-unks} - a parameter efficient approach for cross-lingual transfer leveraging the modularity, and portability of adapters~\citep{Houlsby2019ParameterEfficientTL}. We followed the same \textbf{zero-shot} setup as \citet{alabi-etal-2022-adapting}, however, we make use of \texttt{hau} and \texttt{swa} as source languages since they cover all the news topics used by all languages. 
The setup is as follows: (1) We train language adapters using monolingual news corpora of our focus languages. We perform language adaptation on the \textit{news} corpus to match the domain of our dataset, similar to \cite{alabi-etal-2022-adapting}. (2) We train a task adapter on the source language labelled data using source language adapter. (3) We substitute the source language adapter with the target language to run prediction on the target language test set, while retaining the task adapter. 
    
\paragraph{PET/iPET}~\citep{schick-schutze-2021-exploiting,schick-schutze-2021-just}, also known as (\textbf{I}terative) \textbf{P}attern \textbf{E}xploiting \textbf{T}raining is a semi-supervised approach that makes use of few labelled examples and a prompt/pattern to a LM for few-shot learning. It involves three steps. (1) designing of a prompt/pattern and a verbalizer (that maps each label to a word from LM vocabulary). (2) train an LM on each pattern based on few labelled examples (3) distill the knowledge of the LM on unlabelled data. Therefore, PET leverages unlabelled examples to improve few-shot learning. iPET on the other hand, repeats step 2 and 3 iteratively. We make use of the same set of patterns used for AGNEWS English dataset~\citep{NIPS2015_zhang} provided by the PET/iPET authors. The patterns are (1) $P_1(x) = \_\_\_\_: a, b$ (2) $P_2(x) =  a (\_\_\_\_) b$ (3) $P_3(x) = \_\_\_\_- a b$ (4) $P_4(x) = a b (\_\_\_\_)$ (5) $P_5(x)= \_\_\_\_ News: a b$ (6)  $P_6x)= [Category:\_\_\_\_] a b$, where $a$ is the news headline and $b$ is the news text. In evaluation, we take average over all patterns. 

\paragraph{SetFit}\citep{tunstall-lewis-2022-set-fit} is a few-shot learning framework based on sentence transformer models~\citep{reimers-2019-sentence-bert} like LaBSE following two steps. \textbf{Step 1} fine-tunes the sentence transformer model using a few labelled examples with contrastive learning---where positive examples, are $K$-examples from a class $c$, and negative examples pairs are labelled examples with random labels from other classes. Contrastive learning approach enlarges the size of training data in few-shot scenarios. In \textbf{Step 2}, the fine-tuned sentence transformer model is used to extract rich sentence representation for each labelled example, followed by logistic regression for classification. The advantage of this approach is that it is faster and requires no prompt unlike PET. We use this in both \textbf{zero- and few-shot setting}. For the zero-shot setting, SetFit creates dummy example $N$-times (we set $N=8$, similar to the SetFit paper) like \textbf{``this sentence is \{\}''} where \{\} can be any news topic like ``sports''.

\paragraph{Co:here multilingual sentence transformer} co:here introduced a multilingual embedding model \textit{multilingual-22-12}~\footnote{\url{https://docs.cohere.ai/docs/text-classification-with-classify}}, which supports over a hundred languages, including most of the languages included in .  This is only for the few-shot setting.

\paragraph{OpenAI ChatGPT API\footnote{\url{https://openai.com/blog/chatgpt}}} is an LLM trained on a large chunk of texts to predict the next word like GPT-3~\citep{NEURIPS2020_gpt3}, followed by a set of instructions in a prompt based on human feedback. It leverages Reinforcement Learning from Human Feedback (RLHF), similar to InstructGPT~\citep{Ouyang2022TrainingLM} to make the LLM to interact in a conversational way.  We prompt the OpenAI API based on GPT-3.5 Turbo and GPT-4 to categorize articles into news topics. 
For the prompting, we make use of a simple template from \citet{sanh2022multitask}:  \textit{'Is this a piece of news regarding \{\{``business, entertainment, health, politics, religion, sports or technology''\}\}? \{\{INPUT\}\}'}. We make use of the first 100 tokens of \texttt{headline+text} as \{\{INPUT\}\}. The completion of the LLM can be a single word, a sentence, or multiple sentences. We check if a descriptive word relating to any of the news topics has been predicted. For example, ``economy'', ``economic'', ``finance'' is mapped to ``business'' news. We provide more details on the ChatGPT evaluation in  \autoref{sec:chatgpt}. 

For all few-shot settings, we tried $K$ samples/shots per class where $K=5, 10, 20, 50$.  We make use of LaBSE as the sentence transformer for SetFit, and AfroXLMR-large as the LM for PET.

\subsection{Results}

\begin{table*}[t]
\begin{center}
\footnotesize
\begin{adjustbox}{width=\textwidth,center}

\begin{tabular}{lrrrrrrrrrrrrrrrr|c}
\toprule
\textbf{Model} &  \textbf{amh} & \textbf{eng} & \textbf{fra} & \textbf{hau} & \textbf{ibo} & \textbf{lin} & \textbf{lug}  & \textbf{orm} & \textbf{pcm} & \textbf{run} & \textbf{sna} & \textbf{som} & \textbf{swa} & \textbf{tir} & \textbf{xho} &  \textbf{yor} & \textbf{AVG}  \\
\midrule
\multicolumn{7}{l}{\textit{Fine-tune (AfroXLMR-large)}} \\
5-shots  &68.4&55.1&58.0&35.8&71.3 & 52.7 & 29.2 & 39.2 &92.5 &71.2&70.2& 18.1 &42.5& 30.2&  46.5   &62.7&  52.7\\
10-shots  &75.5&75.2& 65.9& 64.6&86.1&72.6& 31.3 & 56.8 &95.8&87.3&80.8 & 38.9 &73.8& 36.3 &  61.7 & 69.4&  67.0 \\
20-shots  &88.5&85.6&78.3&85.2 &90.4&80.8 & 48.4 &41.1&97.4 &90.0& 92.3 & 63.6 & 82.9& 67.3&  83.1  & 84.3&  78.7 \\
50-shots  &91.4 &87.5&86.9 &88.8&87.3&91.0& 75.2 &71.3&96.4 &89.8&95.5& 85.3 & 86.6& 86.2 &  94.1 & 90.2&  87.7 \\
\midrule
\multicolumn{7}{l}{\textit{Fine-tune (LaBSE)}} \\
5-shots  &71.6&67.4&61.3 &60.7&63.6 & 65.9 & 59.5 &43.3 &86.5 &65.6&83.1& 25.4 & 49.1& 36.1 & 46.0 & 71.2 &  59.7 \\
10-shots  &79.0&77.1&76.8&79.7&77.1&70.2 & 68.3 &58.5 &94.5 &81.9 &84.8 & 44.8 & 77.2&51.8 & 69.9 & 79.8 &  73.2 \\
20-shots  &90.3&84.7&83.1&85.1&82.0&82.2 & 70.4 &72.3& 95.5&86.0 &90.6 & 66.6 & 84.3 & 69.0 & 80.5 & 86.0&  81.8 \\
50-shots  &89.6&86.3&85.6&87.1&86.4 &88.4 & 80.6 &77.8&  96.7&87.9& 93.0 & 80.1 & 85.3& 79.6& 87.4& 88.6&  86.3 \\
\midrule
\multicolumn{7}{l}{\textit{PET}} \\
5-shots  & 89.9& 80.8&72.3& 82.6&85.0& 82.9 &79.0 &89.2 &94.5&87.7 &88.9 & 69.5 & 79.6& 59.7 & 84.3 &84.0&  81.9
 \\
10-shots  & 91.1&81.7&83.3&86.6 &86.1&87.6  &84.0& 91.8&96.6&90.8&91.4 & 74.9 & 81.1 & 69.2 & 88.9 & 90.5&  86.0
\\
20-shots  &92.7&86.4&82.8&89.1&88.6&89.2& 83.8& 94.9&96.7&88.7&93.3 & 81.6 & 83.5 & 72.4 & 91.5 & 91.0&  87.9
 \\
50-shots  &92.9&89.2 &89.1 &90.9&90.6&89.6&86.7 &96.0&97.2& 90.9& 94.8 & 84.2 & 84.2& 76.4 & 93.5 & 92.4 &  89.9
 \\
\midrule

\multicolumn{7}{l}{\textit{SetFit}} \\
5-shots   &  68.3  &  69.6  &  64.3  &  76.0  &  78.9  &  48.3  & 28.9 &  38.8  &  91.2  &  74.8  &  85.8  &  68.9 &  76.8  & 73.1 & 84.0 &  60.2  & 68.0   \\
10-shots  &  84.8  &  82.0  &  80.5  &  79.4  &  71.4  &  77.8  & 49.5 &  57.3  &  92.8  &  83.8  &  89.2  &  65.1 &  81.2  & 64.9 & 83.6 & 76.5  &  76.2 \\
20-shots  &  87.9  &  78.5  &  83.9  &  83.3  &  81.8  &  86.6  & 71.7 &  61.0  &  97.4  &  87.0  &  83.2  & 69.4 &  79.2  & 64.9  & 78.4 & 85.0  & 80.0  \\
50-shots  &  88.6  &  76.6  &  83.8  &  83.0  &  77.3  &  81.9  & 60.8 &  63.6  &  93.6  &  85.6  &  90.6  &  67.9 &  76.5  &  69.8  & 83.8 & 86.0  &  79.3 \\
\midrule
\multicolumn{7}{l}{\textit{Cohere sentence embedding API}} \\ 
5-shots  &66.0 & 65.9 & 60.2 & 74.2 & 72.0 & 69.8 & 50.2 & 50.0 & 74.0 & 61.2 & 78.1 & 52.8 & 67.7 & 60.1 & 68.3  & 71.9 & 65.2   \\
10-shots  &80.1 & 72.5 & 71.4 & 80.4 & 75.7 & 78.4 & 65.5 & 57.2 & 84.9 & 78.2 & 85.0 & 60.4 & 73.8 & 59.8  & 83.2 & 80.1 &  74.2 \\
20-shots  &87.6 & 78.0 & 78.4 & 82.9 & 77.7 & 86.9 & 70.2 & 63.9 & 88.7 & 82.7 & 86.6 & 65.3 & 79.0 & 64.8  & 88.2 & 83.9 &  79.1  \\
50-shots & 90.2 & 80.9 & 83.2 & 85.6 & 81.9 & 87.7 & 78.0 & 70.6 & 94.9 & 84.1 & 90.5 & 68.9 & 77.6 & 72.8 & 90.4  & 88.4 &  82.9 \\


\bottomrule
    \end{tabular}
\end{adjustbox}
\vspace{-3mm}
\caption{\textbf{Few-shot learning on }. We compare several few-shot learning approaches: PET, SetFit and Cohere Embedding API.}
\label{tab:fewshot_result}
  \end{center}
  \vspace{-3mm}
\end{table*}

\subsubsection{Zero-shot evaluation}

\paragraph{GPT-3.5-Turbo performs poorly on non-Latin scripts}
\autoref{tab:zeroshot_result} shows the result of zero-shot evaluation using \textsc{Fine-tune}, \textsc{MAD-X},  \textsc{PET}, \textsc{SetFit} and \textsc{GPT-3.5-Turbo} (March 2023 version). Our result shows that cross-lingual zero-shot transfer from a source language with same domain and task (i.e \textsc{Fine-tune} \& \textsc{MAD-X}), gives superior result ($+11$ F1) than PET, SetFit, and \textsc{GPT-3.5-Turbo}. \textsc{GPT-3.5-Turbo} gave better results with over $+9.0$ F1 point better than \textsc{SetFit} and \textsc{PET} showing that capabilities of instruction-tuned LLMs over smaller LMs.  However, the results of \textsc{ChatGPT} were poor ($<42.0)$ for non-Latin based languages like Amharic and Tigrinya which makes use of the Ge'ez script. The languages that make use of Latin script have over $59.0\%$. Surprisingly, some results of \textsc{GPT-3.5-Turbo} are comparable to the \textsc{Fine-tune} approach  for some languages (English, Luganda, Oromo, Naija, Somali, isiXhosa, and \yoruba),  without leveraging any additional technique apart from prompting the LLM.

\paragraph{GPT-3.5-Turbo evaluation improves with newer versions}
We repeated \textsc{GPT-3.5-Turbo} evaluation using a newer version (May 23, 2023 version), our results suggest a significant improvement of the result for 14 (out of 16) languages in our evaluation. This implies that the newer version of the model seems to be better than older versions for the news topic classification task. 

\paragraph{GPT-4 overcomes the limited non-Latin capabilities of GPT-3.5-Turbo}
We also evaluated on \textsc{GPT-4} on the 16 languages in zero-shot setting. Our results shows a significant improvement in performance over \textsc{GPT-3.5-Turbo} by over $+9$ points. Surprinsingly, GPT-4 was able to overcome the limitation of \textsc{GPT-3.5-Turbo} for languages with non-Latin script (i.e Amharic and Tigrinya) with impressive performance, matching the performance of cross-lingual transfer experiment from a related African language (i.e. \textsc{Fine-tune} \texttt{hau/swa}$\rightarrow$ \texttt{xx} and \textsc{MAX-X} \texttt{hau}$\rightarrow$ \texttt{xx}).

The large performance gap between GPT-3.5-Turbo and GPT-4 may be due to either the former being a distilled version of a more powerful model created to reduce inference cost, which also significantly affected its performance on non-Latin scripts.\footnote{https://arstechnica.com/information-technology/2023/07/is-chatgpt-getting-worse-over-time-study-claims-yes-but-others-arent-sure/}~\footnote{ https://platform.openai.com/docs/models/gpt-3-5}  Alternatively, GPT-4 may just be a bigger and better model with more multilingual and non-Latin capabilities. 

\paragraph{Leveraging labelled data from other languages is more effective}In general, it may be advantageous to consider leveraging knowledge from other languages with available training data when no labelled data is available for the target language. Also, we observe that Swahili (\texttt{swa}) achieves better result as a source language than Hausa (\texttt{hau}) especially when transferring to \texttt{fra} ($+13.8$), \texttt{lug} ($+9.0$), and \texttt{eng} ($+3.6$). The reason for the impressive performance from Swahili to Luganda might be due to both languages belonging to the same Greater Lake Bantu language sub-group, but it is unclear why Hausa gave worse results than Swahili when adapting to English or French.  However, with few examples, PET and SetFit methods are powerful without leveraging training data and models from other languages.

\subsubsection{Few-shot evaluation}
\label{sec:few_shot_eval}
\autoref{tab:fewshot_result} shows the result of the few-shot learning approaches. With only 5-shots, we find all the few-shot approaches to be better than the usual \textsc{fine-tune} baselines for most languages.  However, as the number of shots increases, they have comparable results with \textsc{SetFit} and \textsc{co:here API} especially for $K=20, 50$ shots. However, we found that \textsc{PET} achieved very impressive results with 5-shots ($81.9$ on average), matching the performance of \textsc{SetFit}/\textsc{co:here API} with 50-shots. The results are even better with more shots i.e ($k=10$, $86.0$ F1),  ($k=20$, $87.9$ F1), and  ($k=50$, $89.9$ F1). Surprisingly, with 50-shots, \textsc{PET} gave competitive result to the full-supervised setting (i.e. fine-tuning all TRAIN data) that achieved ($92.6$ F1) (see \autoref{tab:masakhanews_baselines}). It's important to note that PET make use of additional unlabelled data while SetFit and Cohere API do not. In general, our result highlight the importance of getting few labelled examples for a new language we are adapting to, even if it is as little as 10 examples per class---which is typically not time-consuming to annotate
~\citep{lauscher-etal-2020-zero,hedderich-etal-2020-transfer}.

\section{Conclusion}
In this paper, we created the largest news topic classification dataset for 16 typologically diverse languages spoken in Africa. 
We provide an extensive evaluation using both full-supervised and few-shot learning settings. Furthermore, we study different techniques of adapting prompt-based tuning and non-prompt methods of LMs to African languages. Our experimental results shows that prompting LLMs like ChatGPT perform poorly on the simple task of text classification for several under-resourced African languages especially for non-Latin based scripts. Furthermore, we showed the potential of prompt-based few-shot learning approaches like PET (based on smaller LMs) for African languages.  Our work shows that existing supervised approaches work well for all African languages and that language models with only a few supervised samples can reach competitive performance, both findings which demonstrate the applicability of existing NLP techniques for African languages. 

In the future, we plan to extend this dataset to more African languages,  include the evaluation of other multilingual LLMs like BLOOM, mT0~\citep{muennighoff2022crosslingual} and XGLM~\citep{lin-etal-2022-shot}, and extend analysis to other text classification tasks like sentiment classification~\citep{shode2022yosm,shode-etal-2023-nollysenti, muhammad2023afrisenti}.

\section{Limitations}
One major limitation of our work is that we did not evaluate extensively the performance of ChatGPT LLM on several African languages and tasks such as question answering, and text generation tasks. Our evaluation is only limited to text classification and may not generalize to many tasks. However, we feel that if it perform poorly on text classification, the result may even be worse on more difficult NLP tasks. Also, there is a challenge that our result may not be fully reproducible since we use the ChatGPT API where the underlining LLM are often updated or improved with time. It might be that the support for non-Latin based script may improve significantly in few months. This limitation also applied to the co:here embedding API. 

\section{Ethics Statement}
Our work aims to provide benchmark dataset for African languages, we do not see any potential harms when using our news topic classification datasets and models to train ML models, the annotated dataset is based on the news domain, and the articles are publicly available, and we believe the dataset and news topic annotation is unlikely to cause unintended harm. Also, we do not see any privacy risks in using our dataset and models  because it is based on news domain. 
\subsubsection*{Acknowledgments}
We would like to thank Yuxiang Wu for the suggestions on the few-shot experiments. We are grateful for the feedback from the anonymous reviewers of AfricaNLP and IJCNLP-AACL that helped improved this draft. David Adelani acknowledges the support of DeepMind Academic Fellowship programme. This work was supported in part by Oracle Cloud credits and related resources provided by Oracle. Finally, we are grateful to OpenAI for providing API credits through their Researcher Access API programme to Masakhane for the evaluation of GPT-3.5 and GPT-4 large language models.

\bibliography{anthology,custom}
\bibliographystyle{acl_natbib}

\appendix
\section{Annotation Tool}
\label{sec:tool}
\begin{figure*}[ht]
    \centering
    \frame{\includegraphics[width=0.94\linewidth]{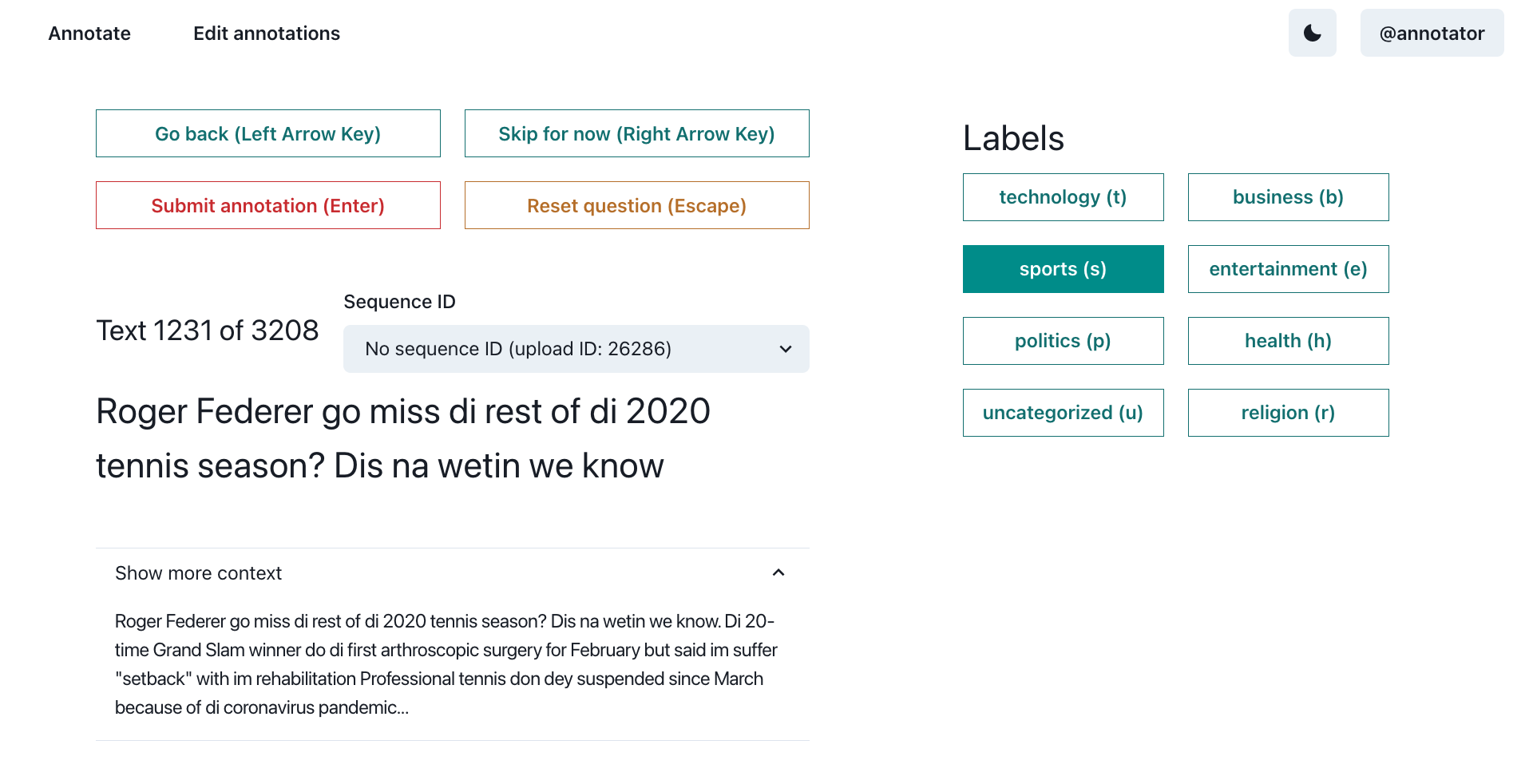}}
    \vspace{-2mm}
    \caption{\textbf{Interface of our in-house Annotation tool}. Annotators can correct the pre-defined category assigned and also edit their annotation}
    \label{fig:annotation_tool}
\end{figure*}

\autoref{fig:annotation_tool} provides an example of the interface of our in-house annotation tool. 

\section{Comparing different article content types} 
\label{sec:article_type}
\autoref{tab:headline_only} provides the comparison between using only news \texttt{headline} and \texttt{headline+text} for training. We find significantly improvement on average when we make use of \texttt{headline+text} for training across all models and languages especially for classical ML methods (MLP, NaiveBayes, and XGBoost). 

\section{ChatGPT Evaluation} 
\label{sec:chatgpt}

We prompted ChatGPT for news topic classification using the following template: \textit{'Is this a piece of news regarding \{\{``business, entertainment, health, politics, religion, sports or technology''\}\}? \{\{INPUT\}\}'}. The completion may take different forms e.g. a single word, sentence or multiple sentences. Examples of such predictions are: 
\begin{enumerate}
    \item sports
    \item This is a piece of news regarding sports.
    \item This is a piece of sports news regarding the CHAN 2021 football tournament in Cameroon. It reports that the Mali national football team has advanced to the semi-finals after defeating the Congo national team in a match that ended in a penalty shootout.
    \item This is a piece of news regarding sports. It talks about the recent match between Tunisia and Angola in the African Cup of Nations. Both teams scored a goal, and the article mentions some of the details of the game, such as the penalty and missed chances.
    \item I'm sorry, but I'm having trouble understanding this piece of news as it appears to be in a language I don't recognize. Can you please provide me with news in English so I can assist you better?
\end{enumerate}

To extract the right category, we make use of a simple verbalizer that maps the news topic to several indicative words (capitalization ignored) for the category like: 
\begin{enumerate}[label=(\alph*)]
    \item 'business': \{'business', 'finance', 'economy'. 'economics' \}
    \item 'entertainment': \{'entertainment' , 'music' \}
    \item 'health': \{'health' \}
    \item 'politics': \{'politics', 'political' \}
    \item 'religion': \{'religion' \}
    \item 'sports': \{'sports', 'sport' \}
    \item 'technology': \{'technology' \}
\end{enumerate}

When the right category is not obvious, like (5 : ``I'm sorry, but I'm having trouble understanding this piece of news as it appears to be in a language I don't recognize. ''), we choose a random category before computing F1-score.

\begin{table*}[t]
 \begin{center}
 \scalebox{0.82}{
 \footnotesize
  \begin{tabular}{lrrrl}
    \toprule
    \textbf{LLM} & \textbf{LLM size} & \textbf{\# Lang.} & \textbf{\# African Lang.} & \textbf{Focus languages covered} \\
    \midrule
    XLM-R-base/large & 270M/550M &100 & 8 &  amh, eng, fra, hau, orm, som, swa, xho  \\
    AfriBERTa-large & 126M & 11 & 11 &   amh, hau, ibo, orm, pcm, run, swa, tir, yor \\
    mDeBERTa & 276M &  110 & 8 & amh, eng, fra, hau, orm, swa, xho   \\
    RemBERT & 575M &  110 & 12 &  amh, eng, fra, hau, ibo, sna, swa, xho, yor   \\
    AfriTeVa-base & 229M & 11 & 11&  amh, run, hau, ibo, orm, pcm, swa, tir, yor \\
    AfroXLMR-base/large & 270M/550M & 20 & 17 & amh, eng, fra, hau, ibo, orm, pcm, run, sna, swa, xho, yor \\
    AfriMT5-base & 580M & 20 & 17 & amh, eng, fra, hau, ibo, orm, pcm, run, sna, swa, xho, yor \\
    FlanT5-base & 580M & 60 & 5 & eng, fra, ibo, swa, yor \\

    \bottomrule
  \end{tabular}
  }
  \caption{Languages covered by different multilingual Models and their sizes }
  \label{tab:plm_languages}
  \end{center}
  \vspace{-3mm}
\end{table*}
              
\begin{table*}[t]
\begin{center}
\footnotesize
\begin{adjustbox}{width=\textwidth,center}

\begin{tabular}{lcrrrrrrrrrrrrrrrr|c}
\toprule
\textbf{Model} & \textbf{size} & \textbf{amh} & \textbf{eng} & \textbf{fra} & \textbf{hau} & \textbf{ibo} & \textbf{lin} & \textbf{lug} & \textbf{orm} & \textbf{pcm} & \textbf{run} & \textbf{sna} & \textbf{som} & \textbf{swa} & \textbf{tir} & \textbf{xho} & \textbf{yor} & \textbf{AVG}  \\
\midrule
\multicolumn{7}{l}{\textit{Headline}} \\
MLP & $<$20K & 86.7 & 72.6 & 69.8 & 80.4 & 77.8 & 79.4 & 74.6 & 81.9 & 87.5 & 73.8 & 84.9 & 71.4 & 69.3 & 80.7 & 79.1 & 83.0 & 78.3 \\
NaiveBayes & $<$20K  & 88.8 & 71.6 & 70.0 & 76.6 & 75.8 & 74.0 & 74.6 & 74.2 & 82.6 & 64.3 & 79.5 & 61.7 & 60.6 & 66.0 & 72.5 & 81.4 & 73.4 \\
XGBoost & $<$20K & 83.6 & 71.3 & 67.8 & 77.4 & 71.3 & 76.7 & 68.7 & 77.7 & 80.8 & 71.3 & 84.6 & 63.4 & 66.4 & 62.1 & 69.4 & 77.5 & 73.1  \\
AfroXLMR-base & 270M & 91.8 & 87.0 & 92.0 & 89.2 & 87.8 & 89.0 & 87.4 & 87.4 & 97.4 & 87.8 & 94.5 & 85.9 & 85.0 & 85.7 & 93.5 & 88.6 & 89.4 \\
AfroXLMR-large & 550M & 93.0 & 89.3 & 91.8 & 91.0 & 90.7 & 91.4 & 87.7 & 90.9 & 98.2 & 89.3 & 95.9 & \textbf{87.1} & 86.6 & 88.5 & 96.2 & 90.3 & 91.1\\
\midrule
\multicolumn{7}{l}{\textit{Headline+Text}} \\
MLP & $<$20K & 92.0 & 88.2 & 84.6 & 86.7 & 80.1 & 84.3 & 82.2 & 86.7 & 93.5 & 85.9 & 92.6 & 71.1 & 77.9 & 81.9 & 94.5 & 89.3 & 85.7 \\
NaiveBayes & $<$20K  & 91.8 & 83.7 & 84.3 & 85.3 & 79.8 & 82.8 & 84.0 & 85.6 & 92.8 & 79.9 & 91.5 & 74.8 & 76.6 & 71.4 & 91.0 & 84.0 & 83.7 \\
XGBoost & $<$20K & 90.1 & 86.0 & 81.2 & 84.7 & 78.6 & 74.8 & 83.8 & 83.2 & 93.3 & 79.2 & 94.3 & 68.5 & 74.9 & 75.2 & 91.1 & 85.2 & 82.8  \\
AfroXLMR-base & 270M & 94.2 & 92.2 & \textbf{92.5} & 91.0 & 90.7 & 93.0 & 89.4& 92.1 & 98.2 & 91.4 & 95.4 & 85.2 & 88.2 & 86.5 & 94.7 & 93.0 & 91.7 \\
AfroXLMR-large & 550M & \textbf{94.4} &\textbf{93.1} & 91.1 & \textbf{92.2} & \textbf{93.4} & \textbf{93.7} & \textbf{89.9} & \textbf{92.1} & \textbf{98.8} & \textbf{92.7} & \textbf{95.4} & 86.9 & \textbf{87.7} & \textbf{89.5} & \textbf{97.3} & \textbf{94.0} & \textbf{92.6}\\
\bottomrule
    \end{tabular}
\end{adjustbox}
\caption{\textbf{Baseline results on }. We compare different article content types (i.e \texttt{headline} and \texttt{headline+text}) used to train news topic classification models. Average is over 5 runs.  Evaluation is based on weighted F1-score. }
\label{tab:headline_only}
  \end{center}
  \vspace{-2mm}
\end{table*}




\end{document}